\documentclass{article}
\usepackage{graphicx}
\usepackage{subfigure}
\usepackage{wrapfig}
\usepackage{cite}
\usepackage{multirow}
\usepackage{setspace}
\usepackage{lineno}

\title{Counting Fish and Dolphins in Sonar Images Using Deep Learning}

\author{Stefan Schneider \\
\scriptsize{School of Computer Science} \\
\scriptsize{University of Guelph} \\
\scriptsize{sschne01@uoguelph.ca}
\and 
Alex Zhuang \\ 
\scriptsize{Department of Computer Science} \\
\scriptsize{University of Toronto} \\
\scriptsize{alexdayouzhuang@gmail.com}}

\begin{document}
\maketitle

\begin{abstract}
Deep learning provides the opportunity to improve upon conflicting reports considering the relationship between the Amazon river's fish and dolphin abundance and reduced canopy cover as a result of deforestation. This topic has received increased attention as a result of increased deforestation efforts and large-scale fires of the Amazon Rainforest. Current methods of fish and dolphin abundance estimates are performed by on-site sampling using visual and capture/release strategies. We propose a novel approach to calculating fish abundance using deep learning for fish and dolphin estimates from sonar images taken from the back of a trolling boat. We consider a data set of 143 images ranging from 0-34 fish, and 0-3 dolphins provided by the Fund Amazonia research group. This data set offers unique challenges as the number of images is too small for traditional deep learning methods. To overcome the data limitation, we test the capabilities of data augmentation on an unconventional 15/85 training/testing split. Using 20 training images, we simulate a gradient of data up to 25,000 images using augmented backgrounds and randomly placed/rotation cropped fish and dolphin taken from the training set. We then train four multitask network architectures: DenseNet201, InceptionNetV2, Xception, and MobileNetV2 to predict fish and dolphin numbers using two function approximation methods: regression and classification. For regression, Densenet201 performed best for fish and Xception best for dolphin with mean squared errors of 2.11 and 0.133 respectively. For classification, InceptionResNetV2 performed best for fish and MobileNetV2 best for dolphins with a mean error of 2.07 and 0.245 respectively. Considering the 123 testing images, our results show the success of data simulation for limited sonar data sets. We find DenseNet201 is able to identify dolphins after approximately 5000 training images, while fish required the full 25,000. Our method can be used to lower costs and expedite the data analysis of fish and dolphin abundance to real-time along the Amazon river and river systems worldwide.
\newline
\end{abstract}

% no keywords

\section{Introduction}
The Amazon River is the world's second longest river, shorter only to the Nile, at 6,400km in length. The water basin for the river is 7,050,000 square kilometers in size, covering approximately 40\% the area of South America, where annual flooding of surrounding forest regions is common. The Amazon River contains 5,600 fish species, which is 10\% of all known vertebrate species on Earth \cite{albert2011historical}. The variability of fish abundance throughout the river system varies relative to a variety of factors such as: the river depth, the speed of the current, oxygen content, and canopy cover \cite{bojsen2002effects, lobon2015importance}. 
As a result of its sheer size, surveying the Amazon River is an enormous task where increased efficiencies would drastically improve conservation research efforts. Currently fish and dolphin abundance surveys are labour intensive, involving fishing nets and manual visual spotting performed on suspended platforms in the center of the river \cite{gomez2012population}. 

Canopy cover has been a topic of additional concern for fish populations due to continued deforestation of the Amazon, and recently, intense wild fires. Multiple studies have been conducted globally comparing canopy cover to fish abundance with contradicting results, however, few studies consider the Amazon River specifically \cite{harding1998stream, jones1999effects}. 

In addition to fish abundance, ecologists and conservationists have gone to great lengths to monitor the endangered populations of the Amazon River dolphins: Amazon River dolphin (aka.~boto) \textit{Inia geoffrensis}, Bolivian River dolphin \textit{Inia geoffrensis boliviensis}, Araguaian boto \textit{Inia araguaiaensis}, and tucuxi \textit(Sotalia fluviatilis), are crucial for the Amazon River's ecosystem stability \cite{layne1958observations, hamilton2001evolution}.

The currently implemented method to evaluate and monitor river dolphin populations involves manning four people on each of 1,000+ line and strip transects covering 2,704 kilometers of the river system as a method to identify habitats critical for dolphins \cite{gomez2012population}. Dolphin abundance data is important as a means of species well-being, but also a measure of increased poaching activity \cite{loch2009conflicts}.  In 2013, Mintzer et al. outlined the effect of illegal harvest on survival of Amazon river dolphins, reporting that the decline of apparent survival exceeds the rate of conservation limits \cite{mintzer2013effect}. The labour intense process of current surveying methods are a limiting factor for data collection and, as a result, real conservation policy action \cite{mintzer2013effect}.

Here we successfully demonstrate a deep learning computer vision technique that modernizes current methods of fish and dolphin abundance data collection. Instead of manual surveys, our method analyzes images taken from the back of a trolling boat using the StarFish Seabed Imaging System \cite{starfishseabed}. Utilizing a custom form of data augmentation, we demonstrate that one can successfully train a deep learning system to estimate fish and dolphin population abundance within +/- 2.07 fish and +/- 0.133 dolphins respectively on limited data. This technique can revolutionize the way fish and dolphin populations are monitored and ultimately improve conservation efforts of the Amazon River fish and dolphin populations.

\begin{figure}
    \centering
	\includegraphics[width=9cm]{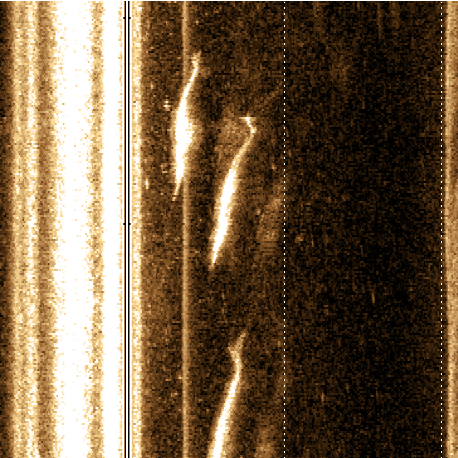}
	\caption{Example image of 3 dolphin testing image.}
\end{figure}

\section{Related Works}

The previous best methods for abundance counts from sonar occur from `dual-frequency identification sonar' (DIDSON) systems. These systems are primarily used for fisheries to estimate the number of individuals that are present in fish farms, fish transfers, or the number of fish that have escaped from the farming operation \cite{holmes2006accuracy, han2009automated}. The system involves mounting the DIDSON camera at a location that has a non-changing background. The system then considers a variety of metrics and thresholds relative to signal recovery to determine fish numbers and error ranges from 1.20 to 13.96 depending on the number of fish in the image \cite{holmes2006accuracy, boswell2008semiautomated}. It has also been shown that DIDSON systems can be used to estimate approximate fish sizes \cite{kang2011semiautomated}. The DIDSON systems are unrealistic for our use-case, however, as they require a fixed location, which is impractical for large scale surveys of river systems. 

While not specific to sonar, computer vision methods for fish species recognition and counting from video data has been performed successfully as well \cite{sung2017vision}. Terayama et al., used Generative Adversarial Networks to transform sonar images from nighttime to daytime for improved results for when quantifying the number of fish in fishery operations \cite{terayama2019integration}.

Our task and images are unique in the realm of computer vision for there has been little to no work on counting the number of fish and dolphins using sonar from the back of a trolling boat. The noise caused by the propeller blade, in addition to the shallow, yet varying depths, create noisy conditions previously not considered for deep learning models.

\section{Background}

Deep learning has seen a rapid growth of interest in many domains, including ecology, due to improved computational power and the availability of large data sets \cite{lecun2015deep, schneider2018deep}. Deep learning methods utilize a multi-layered neural network to solve data representation tasks. Weights within the layers are modified via training, using a form of gradient descent to optimize the a given loss related to desired performance of the model, such as accuracy. Neural networks use a variety of computational techniques including updatable parameters, non-linear transformations, and back-propagation to map logical relationships from input to output \cite{hornik1991approximation}.  In recent years, deep learning methods have dramatically improved performance levels in the fields of speech recognition, object recognition/detection, drug discovery, genomics, and other areas. 

Many recent advances in deep learning have come from improving the architectures of a neural network. One such architecture is the \textit{Convolutional Neural Network} (CNN), which is now the most commonly used architecture for computer vision tasks \cite{fukushima1979neural, krizhevsky2012imagenet}. CNNs introduce convolutional layers within a network which learn \textit{feature maps} representing the spatial similarity of patterns found within the image, such as colour clusters, or the presence or absence of lines \cite{lecun2015deep}. CNNs also introduce \textit{max pooling layers}, a method that reduces computation and increases robustness by evenly dividing these feature maps into regions and returning only their max value \cite{lecun2015deep}.  Many networks have standardized due to their landmark performance including: AlexNet (the first breakthrough CNN), VGG19 (a well-performing 19 layered CNN), GoogLeNet/InceptionNet (which introduced the inception module), ResNet (which introduced residual layers) among many others \cite{krizhevsky2012imagenet, jaderberg2015spatial, szegedy2015going, he2016deep}. One additional feature to networks relative to our sonar quantification is multitask learning, where the model is trained considering multi-labeled outputs \cite{caruana1998multitask}. For our work, utilize multi-task to learning to receive a prediction for both the number of fish and dolphin per image.

Deep learning researchers continually experiment with the modular architectures of neural networks, generally at the trade off of computational cost and memory to accuracy. For our experiment, the models we chose appear on a gradient of increasing complexity: MobileNetV2, NASNetMobile, DenseNet201, Xception, InceptionV3, and Inception-ResNet-V2. Understanding the relative accuracy of these models on ecological images vs. their computational complexity will help map out the classification benefit vs. the computational cost of choosing a particular model.

Performance is limited by the number of labeled images available for training, as the model must be trained on many images in order to produce accurate classifications. A common approach to training deep learning systems on limited data sets is to perform image augmentation. Image augmentation refers to the introduction of variation to an image, such as: mirroring, shifting, rotation, blurring, colour distortion, random cropping, nearest neighbour pixel swapping, among many others \cite{imgaug}. This approach creates new training images, which allows a computer vision network to train on orders of magnitude more examples that uniquely correspond to the provided labeled output classifications. This is a desirable alternative due to the expensive cost (or unavailability) of collecting and labelling additional images. A second common approach to improve performance when training on limited data is by simulating data, where one trains their network on artificially created data to improve model generalization \cite{shrivastava2017learning}. Both these techniques are used in our work. 

\section{Methods}

Our experiment uses a data set of 143 labeled images collected by the Fund Amazonia research group in collaboration with the Operation Wallacea expedition group using the StarFish Seabed Imaging System \cite{fundamazonia, opwall, starfishseabed}. The images are constructed in real time by a side-scan sonar that is able to scan a sweeping angle of 60 degrees within the water. As echoes return, pixel brightness corresponds to the intensity of the echoes reflected, while distance away from the centre line corresponds to the time delayed in the returned echoes. Fish are generally characterized as bright white oval spots and dolphins a silhouette of their outline (Figure 3.1 \& 3.2). A difficulty of this data involves the lighter in appearance oval shapes, similar to fish, but caused by bubbles from the propeller or nearby debris, classified correctly only by experts. Here we utilize a multitask learning framework to quantify the number of fish and dolphin within the images as two outputs of the model. 

\begin{figure}
    \centering
	\includegraphics[width=9cm]{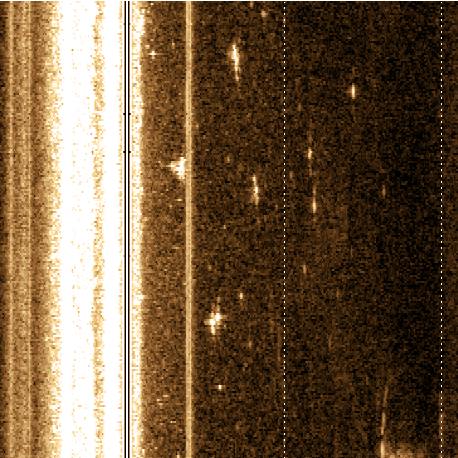}
	\caption{Example image of 10 fish of which DenseNet201 predicts correctly}
\end{figure}

In order to achieve adequate performance using deep learning methods, one requires often thousands-millions of images. In order to satisfy this data requirement we performed an unconventional 15/85 training/test split. Using 20 training images, we removed the cropped and stored the 24 fish and 9 dolphins present to create 20 empty backgrounds. Using blurring, mirroring, and affining, we randomly augmented the 20 backgrounds to create 25,000 empty canvases. We then pasted a uniform random number between 0-34 and 0-3 (the known maximum labels) of fish and dolphin respectively at random locations, augmented their appearance by size and rotation (Figure 3.3). The completed generated image was then augmented using colour changes, blurring, grayscale, dropout, and brightening/darkening. The 25,000 simulated images were split into a 90/10 training/validation split for hyperparameter tuning, and the results reported results on the remaining 123 unseen testing images. Image were resized to 224 * 224 in size, and pixel values were normalized between 0 and 1 as were number of fish considering a regression. For classification, there were 35 classes for fish and 4 for dolphin, corresponding to the number found within the image. Simulated/augmented data operations were performed using the ImgAug library \cite{imgaug}. 

\begin{figure}
    \centering
	\includegraphics[width=9cm]{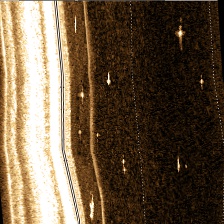}
	\caption{Example simulated example used for training}
\end{figure}

For this experiment, we compare four modern convolutional neural network architectures: DenseNet201, Inception-ResNet-V2, Xception, and MobileNetV2 on the multitask output of fish and dolphin numbers in terms of two statistical function approximators: regression and classification. Due to the variability in simulating the data, we repeat the experiment three times and report the mean error across the 123 testing images considering the number of fish and dolphin per images. In addition, we consider a gradient of performance for the regression DenseNet201 model on fish and dolphin abundance by retraining multiple times considering variable numbers of synthetic images. We represent this relationship utilizing a R\textsuperscript{2} score considering a logarithmic regression. For all experiments, the weights of the network were initialized using Xavier initialization and each model was trained using the Adam optimizer for 150 epochs \cite{he2015delving, kingma2014adam}. 

For the primary experiment of 25,000 augmented images, we found that each network was capable of approximating the number of fish and dolphins. Considering regression: DenseNet201 and Xception performed best for fish and dolphin respectively with 2.11 and 0.133 mean error respectively. Considering classification: InceptionResNetV2 and MobileNetV2 performed best with 2.07 and 0.246 mean error for fish and dolphin respectively (Table 3.1). 

Considering the gradient of performance relative to synthetic images, we find that performance for dolphin abundance plateaued at approximately 5000 images with a logarithmic regression R\textsuperscript{2} value of 0.8554. Considering fish abundance, the logarithmic regression R\textsuperscript{2} value plateaued between approximately 17,500 and 25,000 synthetic images with a value of 0.7334 (Figure 3.4 \& 3.5). 

\begin{figure}[h]
    \centering
	\includegraphics[width=9cm]{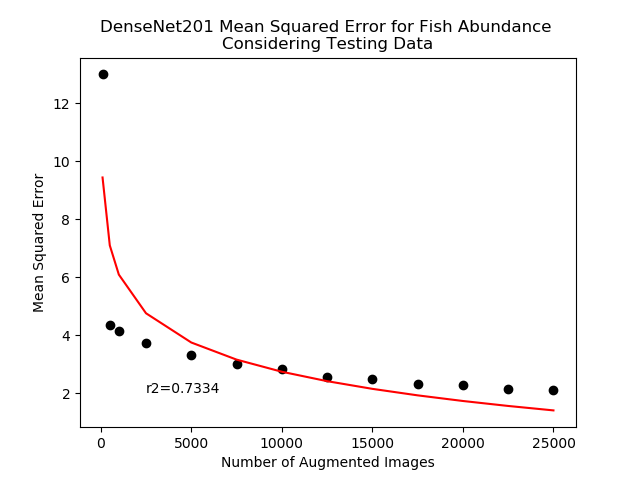}
	\caption{Gradient of mean squared error considering number of synthetic images for fish abundance}
\end{figure}

\begin{figure}[h]
    \centering
	\includegraphics[width=9cm]{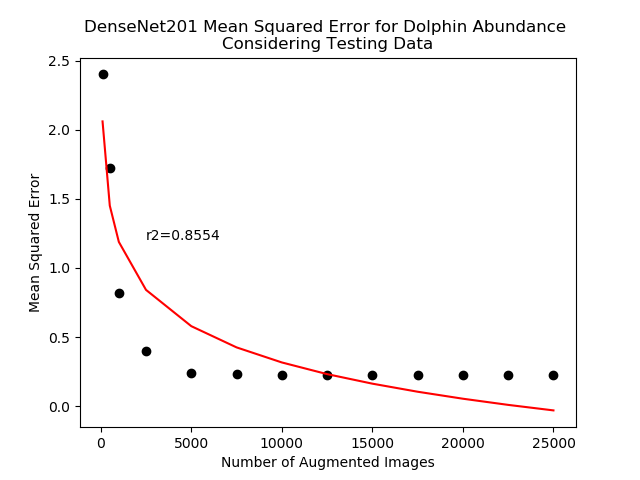}
	\caption{Gradient of mean squared error considering number of synthetic images for dolphin abundance}
\end{figure}

\section{Discussion}
By utilizing modern approaches for computer vision, we demonstrate that researchers can quantify the number of fish and dolphins found in sonar images taken from the back of a trolling boat. Previous studies have demonstrated counting from static sonar images in open water as well as from standard cameras using under water video, but none have demonstrated the quantification of fish and dolphins from variable riverbed backgrounds within the novel river domain. This has applications in the Amazon River, but also in river systems worldwide.

Our results can have a large implication for the data collection methodology of fish systems within river systems and also for conservation efforts of the dolphins of the Amazon River. Our approach allows researchers to quickly collect and analyze data of fish and dolphin abundance, a useful when calculating population metrics. This is particularly important for monitoring endangered dolphin species, which is currently performed by visual call-outs by people on mounted transect locations. Our approach is far less expensive, dangerous, and reliable for all involved. 

Despite a minuscule training data set, we demonstrate the success of deep learning methods when utilizing intelligent data simulation methods. Ecological data is often limited in scope, and methods for synthetic data will be imperative for the use of deep learning methods considering future tasks.

Considering our synthetic data pipeline, we identify that each network is capable of learning the fish/dolphin representation, with Xception performing best on average considering the tasks, and MobileNetV2 the worst. While we tested both regression and classification here, we recommend in practice using regression because it can extrapolate to numbers of fish not seen during training. 

Considering the gradient of performance for the number of synthetic images using DenseNet201, we find that quantifying the number of dolphin silhouettes was ultimately an easier pixel representation task to learn than the fish, requiring less synthetic data than that of the fish silhouettes. Intuitively, this relates to the dolphin pixel representation being more distinct than a fish. In addition, fish abundance labels are likely imperfect due to labelling considering interpretations of what is a fish, a bubble, or debris.  

While we tested counting for fish and dolphins, this methodology can be extended to other applications of sonar images so long as the desired object has a distinct shape. This may include other creatures such as: manta rays, alligators, etc., and in certain cases one may be able to estimate species and/or age based on the size of outline. One may also consider this technique for non-creatures with distinct shapes, such as plant life or improperly discarded objects.

In the wake of increasing numbers of wildfires in the Amazon and in the context of a changing global climate, it has become imperative for research groups and government organizations to understand the extent of damage done to ecosystems. As climate change escalates, biologists need to be able to react to a changing environment equipped with quicker, more efficient analytical tools to do their best work. Deep learning, as demonstrated here, should be one of these tools.

Our hope is that this is just the beginning of a revolution where deep learning improves upon the methods of ecological data collection and analysis. We believe that deep learning can aid in increased accuracies and response times of data collection methods, which can improve the speed at which we recognize species population declines, and increase the speed at which policy decisions are made. 

\section{Conclusion}
Recent advancements in the field of computer vision and deep learning have given rise to reliable methods of feature extraction for regression and classification tasks. The response of Amazon River aquatic populations to continued deforestation and wildfires is generally unknown. In attempt to improve data collection, one proposed method quantifies the number of fish and dolphin individuals from sonar imaging passed through data augmentation. We test the ability of four deep learning architectures: DenseNet201, InceptionResNetV2, Xception, and MobileNetV2 on their ability to quantify the number of fish and dolphin considering images taken from the back of a trolling boat when amplified with synthetic data. We demonstrate successful results quantifying the number of fish and dolphins from sonar images with 2.11 and 0.133 mean error for fish and dolphins respectively considering 123 testing images and quantify a gradient of performance to amount of synthetic data. By adopting our technique, ecologists can rapidly increase their sampling efforts of fish and dolphin abundance in the Amazon River and worldwide. 

\section{Acknowledgments}
We would like the thank the members of Operation Wallacea and Fund Amazonia for their ambition and effort in this work. Specifically Richard Bodmer, Maria Nino, Frederico Barroso, and Kathy Slater. We would also like to acknowledge Stefan Kremer for his supervision and Graham Taylor for their machine learning insights. 

\newpage 

\begin{table}[h]
\centering
\caption{Comparison of Regression and Classification Abundance Predictions for Fish and Dolphins Considering Four Network Architectures}
\begin{tabular}{ c c c c }
	\hline
	\textbf{Animal} & 
	\textbf{Model} & 
	\multicolumn{1}{|p{3cm}|}{\textbf{Regression Mean Error}} &
	\multicolumn{1}{|p{3cm}|}{\textbf{Classification Mean Error}}\\
	\hline
	Fish & DenseNet201 & 2.11 & 2.55 \\
	 & InceptionResNetV2 & 2.514 & 2.07 \\
	 & Xception & 2.27 & 2.31 \\
	 & MobileNetV2 & 2.87 & 2.58 \\
	 \hline
	Dolphin & DenseNet201 & 0.225 & 0.373 \\
	 & InceptionResNetV2 & 0.204 & 0.415 \\
	 & Xception & 0.133 & 0.320 \\
	 & MobileNetV2 & 0.303 & 0.246 \\
	\hline
\end{tabular}
\begin{flushleft}
\end{flushleft}
\end{table}

\newpage
\bibliographystyle{IEEEtran}
\bibliography{MasterReference}

\end{document}